\documentclass[11pt]{article}

\PassOptionsToPackage{dvipsnames,table}{xcolor}

\usepackage[preprint]{acl}

\usepackage{times}
\usepackage{latexsym}

\usepackage{tikz}
\usetikzlibrary{positioning}  %

\usepackage{multirow}

\usepackage[T1]{fontenc}

\usepackage[utf8]{inputenc}

\usepackage[english]{babel}
\usepackage{microtype}

\usepackage{inconsolata}

\usepackage{graphicx}

\usepackage{booktabs}
\usepackage{bm}

\usepackage{subcaption}   %
\usepackage{tcolorbox} %

\newcommand{\az}[1]{\color{black} #1}
\usepackage{xspace}
\newcommand{\BENCHMARK}{\textsc{MDU}\xspace}  
\newcommand{\DATASETSIZE}{23{,}146\xspace}   %
\newcommand{\NTASKS}{13\xspace}       %
\newcommand{\NEVALTASKS}{13\xspace}
\newcommand{\NTASKCATEGORIES}{7\xspace}       
\newcommand{\NMODELS}{45\xspace}      %

\newif\ifwithappendix
\withappendixtrue

\tcbset{
  qabox/.style={
    colback=cyan!10,      %
    colframe=cyan!10,      %
    arc=4mm,               %
    boxrule=0pt,
    width=\linewidth,
    left=4pt, right=4pt, top=4pt, bottom=4pt,
    fontupper=\scriptsize
  }
}

\title{Benchmarking Multilingual Document Understanding in Vision-Language Models}

\author{
  \textbf{Matteo Gabburo\textsuperscript{2}}\thanks{Work done while at Villanova as an external consultant.},
  \textbf{Daniela Baiamonte\textsuperscript{1}},
  \textbf{Stefano Simonazzi\textsuperscript{1}},
\\
  \textbf{Silvia Gardini\textsuperscript{1}},
  \textbf{Sean Cesare Lazzeri\textsuperscript{1}},
  \textbf{Andrea Zugarini\textsuperscript{1}}
\\
  \textsuperscript{1}Villanova,
  \textsuperscript{2}Aithlas
\\
  \small{\texttt{\{dbaiamonte, ssimonazzi, sgardini, slazzeri, azugarini\}@villanova.ai}},
  \small{\texttt{matteo.gabburo@aithlas.com}}
}

\newcommand{\revdel}[1]{}

\newif\ifnotes
\notesfalse   %

\begin{document}
\maketitle
\begin{abstract}
{\az Recent Vision-Language Models (VLMs) perform strongly on multimodal tasks, Document Understanding (DU) among them, but the benchmarks that measure them mostly cover English documents and narrow problems. How well current VLMs generalize across languages, tasks, styles and layouts therefore remains unclear. We introduce \BENCHMARK, a multilingual DU benchmark of \DATASETSIZE examples spanning \NTASKS tasks and five languages: English, French, German, Italian, and Spanish. The tasks isolate complementary capabilities, from surface-level text and Markdown extraction through summarization, classification and named-entity recognition to layout-sensitive passage and layout extraction. We evaluate \NMODELS state-of-the-art open-weight VLMs from 1.7B to 235B parameters, plus one commercial system under the same protocol. The benchmark is far from saturated: the best average is 69.0, only eight of the \NMODELS\ evaluated systems reach 65, and the two best results come from 27B models rather than from the largest systems we evaluate. Performance is uneven in ways that scale does not fix. The gap between the strongest and weakest language does not close as models grow, and it survives controlling for the domain composition of each language. Capabilities also do not improve together: once we partial out overall model quality, per-task scores are largely uncorrelated, and transcription ability in particular trades off against semantic extraction. }
\end{abstract}

\newcommand{\mduchip}[2]{\colorbox{#1!20!white}{\tiny\strut #2}}
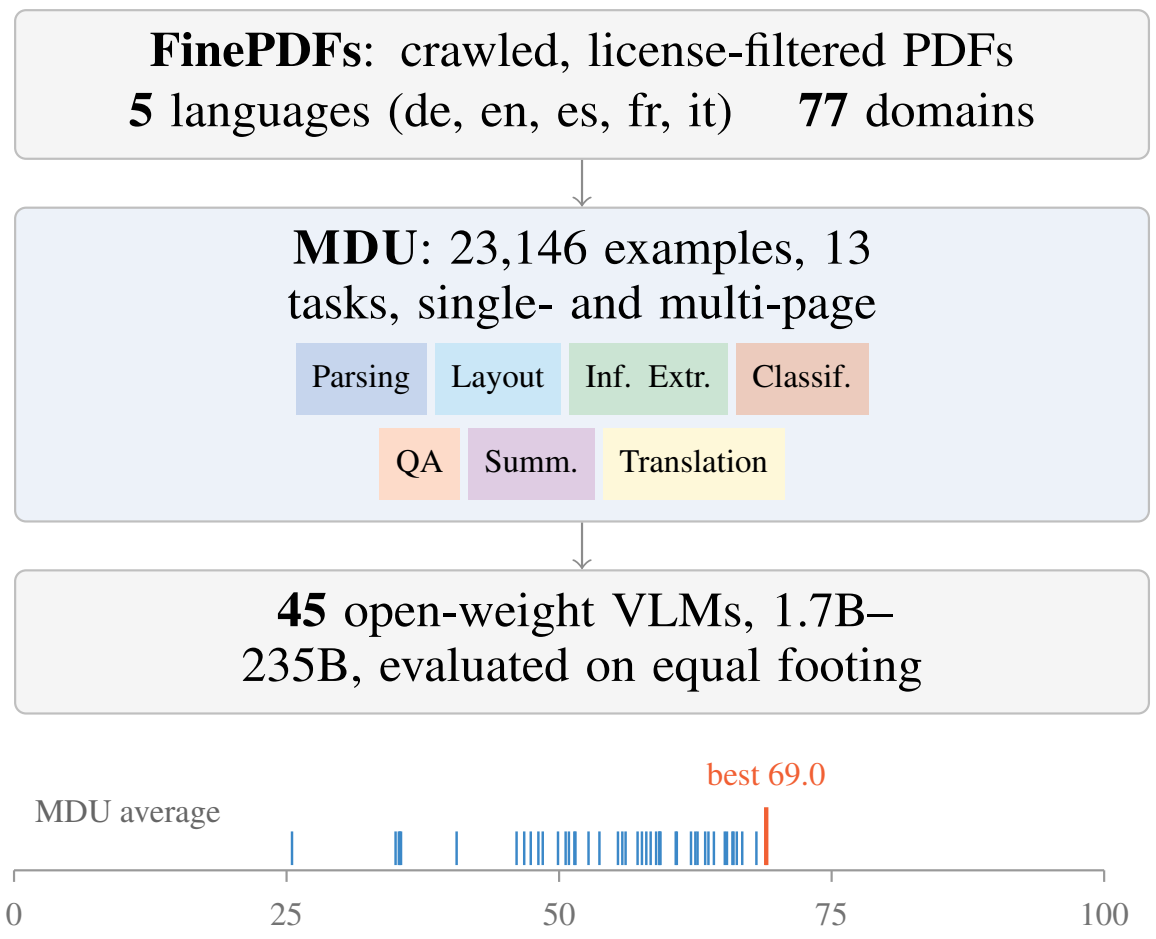
\begin{figure}[t]
\centering
\resizebox{\columnwidth}{!}{%
\begin{tikzpicture}[
  font=\footnotesize,
  band/.style={draw=black!25, rounded corners=2pt, inner sep=4pt, text width=7.0cm,
               align=center},
]
\node[band, fill=black!4] (src) {%
  \textbf{FinePDFs}: crawled, license-filtered PDFs\\[1pt]
  \textbf{5} languages (de, en, es, fr, it) \quad \textbf{77} domains%
};

\node[band, fill=RoyalBlue!6, below=3mm of src] (bench) {%
  \textbf{\BENCHMARK}: \DATASETSIZE\ examples, \NTASKS\ tasks, single- and multi-page\\[2.5pt]
  \mduchip{RoyalBlue}{Parsing}\,\mduchip{Cerulean}{Layout}\,%
  \mduchip{ForestGreen}{Inf.\ Extr.}\,\mduchip{Mahogany}{Classif.}\\[1.5pt]
  \mduchip{RedOrange}{QA}\,\mduchip{Plum}{Summ.}\,\mduchip{Goldenrod}{Translation}%
};

\node[band, fill=black!4, below=3mm of bench] (eval) {%
  \textbf{\NMODELS} open-weight VLMs, 1.7B--235B, evaluated on equal footing%
};

\draw[->, black!45] (src) -- (bench);
\draw[->, black!45] (bench) -- (eval);

\coordinate (rug) at ([yshift=-10mm]eval.south west);
\begin{scope}[shift={(rug)}, x=0.0700cm, y=1.35cm]
  \draw[black!40] (0,0) -- (100,0);
  \foreach \t in {0,25,50,75,100} {%
    \draw[black!40] (\t,0) -- (\t,-0.055);
    \node[below=0pt, font=\tiny, text=black!55] at (\t,-0.055) {\t};
  }
  \foreach \s in {25.5,35.0,35.3,35.4,35.5,40.6,46.1,46.8,47.4,48.1,48.5,49.9,50.6,50.9,51.4,51.5,52.7,53.7,55.4,55.8,56.1,57.2,57.6,58.0,58.4,58.9,59.2,59.2,59.3,60.7,60.8,62.1,62.5,62.7,63.4,63.7,64.2,65.2,65.4,65.9,66.0,66.3,66.8,68.1,69.0}{%
    \draw[RoyalBlue!75, line width=0.45pt] (\s,0.025) -- (\s,0.185);}
  \draw[RedOrange, line width=0.8pt] (69,0.025) -- (69,0.30);
  \node[font=\tiny, text=RedOrange, anchor=south] at (69,0.30) {best 69.0};
  \node[font=\tiny, text=black!60, anchor=west] at (0,0.265) {MDU average};
\end{scope}
\end{tikzpicture}}
\caption{\BENCHMARK\ at a glance. Documents are drawn from license-filtered
FinePDFs across five languages and 77 domains; \NTASKS\ tasks, grouped into seven capability families, are generated over them in both single- and multi-page settings. Each tick on the scale is one of the \NMODELS\ evaluated systems: the best average is 69.0 and only eight of them reach 65, so the suite is far from saturated.}
\label{fig:teaser}
\end{figure}

\section{Introduction}

{\az Document Understanding (DU) has traditionally relied on modular pipelines that chain Optical Character Recognition (OCR), layout analysis, and task-specific models. Each stage adds its own error modes, and the brittle interfaces between them make such systems hard to adapt to new document types, languages or tasks. Recent Vision-Language Models (VLMs) change this: given a document image and a natural-language instruction, one end-to-end model transcribes, summarizes, classifies, extracts and answers questions about a page in a single interface. }

{\az Benchmarks, however, are predominantly in English, and the few multilingual resources are typically confined to a single narrow setting such as form understanding~\cite{xu2022xfund}. Layout, style, entity distributions and format all vary across languages, so English-only evaluation overstates real-world usability. Existing benchmarks also lean heavily on visual question answering and short-answer extraction, which conflates distinct abilities under one format, while practical workflows demand operations that do not necessarily improve together. }

{\az
We address these gaps with \BENCHMARK, a multilingual DU benchmark of \DATASETSIZE examples spanning \NTASKS tasks in five languages: English, French, German, Italian, and Spanish. %
Tasks target complementary capabilities, from document transcription to layout-grounded spatial reasoning, on a heterogeneous set of documents varying in domain, style, and format; roughly 13\% are multi-page, which lets us probe coherence across page boundaries. }

{\az In summary, our contributions are as follows:
\begin{itemize}
    \item We introduce and release \BENCHMARK, a multilingual DU benchmark of \DATASETSIZE examples spanning \NTASKS tasks in five languages, with a task suite that goes beyond visual question answering and an evaluation protocol suited to each task.
    \item We evaluate \NMODELS VLMs from 1.7B to 235B parameters and find that scale explains only part of the variance: model family accounts for more of it than parameter count, and the cross-lingual gap persists at every scale.
    \item We show that the suite measures separable skills: controlling for general model quality, mean partial correlation between tasks drops from $+0.66$ to $+0.09$, while tasks within the same capability family remain correlated ($p=0.009$).
    \item We quantify the self-preference risk inherent in synthetic benchmark construction, and show it concentrates in reference-scored metrics, shrinking threefold under format-independent ones.
\end{itemize}
}

\section{Related Work}

\paragraph{Multilingual document understanding.}
Multilingual resources for document and text-in-image understanding remain scarce, and are narrow either in task coverage or in linguistic scope. MTVQA \citep{tang2025mtvqa} spans nine languages, including French, German and Italian, but a single task and no Spanish. XFUND \citep{xu2022xfund} and its extension SRFUND \citep{ma2024srfund} reach seven and eight languages for form understanding, from key-value extraction to document-level structure recovery, yet stay within that domain and never reach open-ended tasks such as summarization or free-form question answering. CC-OCR \citep{yang2025cc} offers broad linguistic coverage for text recognition alone. DocAtlas \citep{heakl2026docatlas} is the most linguistically extensive resource to date, spanning 82 languages, but its subtasks are parsing-oriented rather than semantic. SEA-Vision \citep{yue2026sea} joins document and scene-text understanding yet is restricted to Southeast Asian languages; MDPBench \citep{li2026mdpbench} covers 17 languages, again for parsing; and HW-MLVQA \citep{pal2025hw} and OCRBench v2 \citep{fu2026ocrbench} are each confined to one language pair and one task.

\paragraph{Task-specific document benchmarks.}
A second line of work isolates individual capabilities that \BENCHMARK\ evaluates jointly. PosterSum \citep{saxena2025postersum} targets abstractive summarization of scientific posters, where the reference is a discursive synthesis rather than a span, the property that motivates our surface-overlap metrics for summarization. TABVERSE \citep{ahsan2026tabversebenchmarkingcrossformattable} benchmarks table understanding across formats, the capability our Table QA task probes within full page images. Both are English-only and single-capability, hence complementary to a multi-task multilingual suite.

\paragraph{English-centric document understanding.}
The benchmarks that shaped the field are overwhelmingly English, and they defined the prevailing notion of document understanding. DocVQA \citep{mathew2021docvqa} established extractive question answering over scanned pages as the canonical task; InfographicVQA \citep{mathew2022infographicvqa} extended it to infographics, where answering requires reading layout and performing elementary numerical reasoning; ChartQA \citep{masry2022chartqa} isolated reasoning over chart imagery; and DUDE \citep{vanlandeghem2023dude} broadened the setting along three axes at once (multiple pages, multiple document types, and questions that admit no answer), making it the closest antecedent of the present work. The models we benchmark were largely developed against these resources, yet none can establish whether the capabilities they measure survive a change of language: each is monolingual, so competence on them is confounded with competence in English. \BENCHMARK\ inherits their task framing and asks the question they cannot pose.

\paragraph{Positioning.}
Existing multilingual resources trade task breadth for linguistic coverage, or the reverse: those broadest in languages stay within parsing and recognition \citep{heakl2026docatlas,li2026mdpbench,yang2025cc}, while those probing richer capabilities confine themselves to one task family \citep{xu2022xfund,tang2025mtvqa} or to English \citep{mathew2021docvqa,vanlandeghem2023dude}. \BENCHMARK\ occupies the missing quadrant, with \NTASKS\ tasks spanning transcription, extraction, classification, summarization, translation and layout-grounded reasoning, evaluated on equal footing across five languages and both single- and multi-page settings.

\section{\BENCHMARK Dataset}
{\az We introduce \BENCHMARK, a novel multilingual dataset to benchmark VLMs in Document Understanding. \BENCHMARK consists of \DATASETSIZE examples, distributed across \NTASKS tasks and five languages: English, French, Italian, German, and Spanish. Tasks were designed to assess different capabilities in DU on a heterogeneous set of documents, varying in language, domain, style and format. }

\subsection{Dataset Creation}
\label{ssec:dataset_creation}
{\az Given a collection of documents $\mathcal{D}$ and a set of pre-defined tasks $\mathcal{T}$, for every task $\mathcal{T}_k$ we build examples as $(\mathbf{x},\mathbf{y}, \mathcal{I}_s)$ triples, where $\mathbf{x}$ is the instruction, $\mathbf{y}$ the expected response and $\mathcal{I}_s$ a list of one or more images, each a rendered document page. We first select documents suitable for $\mathcal{T}_k$; for each $d_i \in \mathcal{D}_k$ we sample one or more pages $p_{i,j} \in d_i$ and render them as images; we then generate a synthetic instruction--response pair grounded on $\mathcal{I}_{i}$, using open-weight VLMs, downstream models and document metadata. Each task-specific approach is validated by human annotators and refined until it reaches a predefined accuracy. }

\paragraph{Documents Selection.}
{\az We build on FinePDFs~\cite{kydlicek2025finepdfs}, a corpus of about 455 million documents in 1733 languages that provides extracted text and metadata (per-page and document language, length, page boundaries, quality scores and source URL) but not the original PDFs. We therefore sampled about 2.5 million documents uniformly across the five target languages and crawled them; crawling failed for about 35\%, leaving 1.6 million.

To support responsible use, we determine the licensing status of each source PDF by pattern matching the document URL for an explicit license statement, extending the search to the hosting website's terms-of-use and legal-notice pages when the document itself carries none. We also record URL reachability and whether the site's \texttt{robots.txt} restricts crawling or AI-related use, and exclude every document with a non-permissive license or \texttt{robots.txt}. Only $2.8\%$ carry an explicitly permissive license, leaving about 44 thousand documents, which further shrinks to 20,770 once documents unsuitable for a given task are discarded; Figure QA, for instance, requires pages with figures or charts large enough to support a question. For the tasks and languages left underrepresented we added 5,444 examples sampled from documents with permissive \texttt{robots.txt} but unspecified license terms. }

\paragraph{Page Sampling.}
{\az To maximize diversity of tasks, languages, styles and source documents we generate one example per document. About 87\% of examples are single-page; the remaining 13\% are multi-page, with the number of pages sampled between 2 and 5.}

\paragraph{Instruction-Response Generation.}
{\az Instructions and responses are grounded in the input image(s). Unless otherwise specified we used \textsc{Qwen/Qwen3.5-122B-A10B}~\cite{qwen3.5}, one of the strongest permissively licensed open-weight models, to create the synthetic examples. Each task has its own pipeline, in the simplest case a task-specific seeding prompt translated across the five languages. To diversify style we follow PersonaHub~\cite{ge2406scaling}, injecting language-specific persona descriptions into the task prompt. }

\paragraph{Validation and Iterative Refinement.}
Native or near-native linguists validated examples in all five languages before large-scale generation. A synthetic pipeline must have the quality of its references measured rather than assumed, so we audited it after generation: for every task except Translation, in every language, annotators rated 50 instruction--response pairs on a four-point scale (\textsc{a}: correct and complete; \textsc{b}: acceptable with minor imperfections; \textsc{c}: substantial errors; \textsc{d}: unusable) against the page images, for 3{,}000 judgements; Translation was audited by translation direction instead. Each judgement also flags whether a failure originated on the instruction or the response side.

Annotators rate every audited task \textsc{a} or \textsc{b} in at least 90\% of cases, median 96\%. These numbers make the rest of the evaluation interpretable: when a model scores poorly here, a human reader had judged its reference sound.

\subsection{Tasks}
\label{ssec:tasks}
\BENCHMARK\ covers \NTASKS\ tasks grouped into \NTASKCATEGORIES\ categories. Table~\ref{tab:tasks} lists each task with the metric it is scored by, the share of its references a human annotator accepted, and its size in every language. Each task has a single-page form and, where the content supports it, a multi-page form over a sequence of contiguous pages; we report the two together under one metric. Appendix~\ref{app:generation} describes how each task is generated, and Appendix~\ref{app:task_examples} shows one example per task.

\paragraph{Parsing.} Text Extraction asks for the visible text of the page, with tables
rendered in GitHub-Flavored Markdown and untranscribable graphics omitted. Markdown Extraction asks instead for the page's hierarchical structure (headings, lists, tables, emphasis) as valid Markdown. Both admit a sentinel response for pages that cannot be faithfully represented in text, keeping unanswerable pages out of the scored set.

\paragraph{Layout extraction.} The model names the elements of a page (headers, footers, tables, figures, text blocks) and localises them with coordinates normalised to $[0,1]$. Requests vary in scope (one element type, several, or all) and in answer format (prose, JSON, or both), and include negative cases where the requested element is absent. \S\ref{ssec:results_and_discussion} explains why we score this task on the page words a box covers rather than on the geometry of the box itself.

\paragraph{Information extraction.} NE Extraction asks for a flat JSON object mapping
each entity mention to a category over a broad, deliberately non-exhaustive taxonomy; mentions are raw spans of the visible text, with no normalisation or linking, which keeps the task at span level. Passage Extraction asks for the segment of the page answering a query, quoted verbatim inside a discursive answer rather than as a bare span.

\paragraph{Classification.} Type Classification assigns a document type and Domain
Classification a subject domain, the latter in a JSON response carrying a confidence level and a short justification. Their instructions enumerate the admissible labels, ten for each single-page variant and nineteen types or fifteen domains for the multi-page ones, so they also measure whether a model respects a prescribed output format.

\paragraph{Question answering.} Three tasks share a response format and differ in what
grounds the question: QA on the running text, Table QA on tabular content, Figure QA on figures and charts. The multi-page variant of QA requires combining evidence scattered across a page sequence rather than locating it on one page. We generate negative cases deliberately (absent content, wrong references, answers that are simply ``no''), so a model cannot score well by always asserting.

\paragraph{Summarization.} Summarization takes an unconstrained request over a page or a
document. Conditional Summarization adds an explicit constraint on the output, drawn from fourteen alternatives covering length, number of list items or table rows, register and style (Appendix~\ref{app:condsum}). The constraint is verifiable on the response alone, so we score the task by constraint satisfaction rather than against a reference.

\paragraph{Translation.} The model renders the whole page into a target language, headings and boilerplate included, rather than a curated passage. Examples cover every ordered pair of the five languages.

\begin{table*}[t]
\centering\small\setlength{\tabcolsep}{4pt}
\begin{tabular}{@{}llll rrrrr r@{}}
\toprule
\multirow{2}{*}{\textbf{Task}} & \multirow{2}{*}{\textbf{Category}} &
\multirow{2}{*}{\textbf{Metric}} & \multirow{2}{*}{\textbf{A\,+\,B}} &
\multicolumn{5}{c}{\textbf{Examples per language}} & \multirow{2}{*}{\textbf{Total}} \\
\cmidrule(lr){5-9}
 & & & & \textbf{de} & \textbf{en} & \textbf{es} & \textbf{fr} & \textbf{it} & \\
\midrule
Text Extraction        & Parsing      & $\mathrm{CER}_{\mathrm{sim}}$          & 97.6 & 400 & 481 & 546 & 511 & 327  & 2{,}265 \\
Markdown Extraction    & Parsing      & $\mathrm{CER}_{\mathrm{sim}}$ (norm.)  & 97.2 & 400 & 365 & 537 & 553 & 389  & 2{,}244 \\
Layout Extraction      & Layout       & $F_1$ over covered words               & 96.0 & 400 & 639 & 380 & 322 & 589  & 2{,}330 \\
NE Extraction          & Inf.\ Extr.  & value-set micro-$F_1$                  & 94.0 & 400 & 400 & 400 & 400 & 1{,}368 & 2{,}968 \\
Passage Extraction     & Inf.\ Extr.  & chrF                                   & 98.4 & 96  & 98  & 103 & 108 & 95   & 500 \\
Type Classification    & Classif.     & label accuracy                         & 96.4 & 88  & 121 & 103 & 100 & 88   & 500 \\
Domain Classification  & Classif.     & label accuracy                         & 99.2 & 91  & 119 & 112 & 87  & 91   & 500 \\
QA                     & QA           & token recall                           & 95.6 & 400 & 563 & 400 & 400 & 937  & 2{,}700 \\
Table QA               & QA           & numeric EM (tol.)                      & 97.2 & 400 & 419 & 630 & 431 & 438  & 2{,}318 \\
Figure QA              & QA           & numeric EM (tol.)                      & 90.8 & 193 & 231 & 163 & 407 & 193  & 1{,}187 \\
Summarization          & Summ.        & chrF                                   & 96.0 & 400 & 489 & 400 & 374 & 911  & 2{,}574 \\
Conditional Summ.      & Summ.        & constraint sat.\ rate                  & 93.2 & 119 & 147 & 158 & 142 & 114  & 680 \\
Translation            & Translation  & chrF++                                 & $^{\dagger}$ & 400 & 400 & 604 & 553 & 423  & 2{,}380 \\
\midrule
\textbf{Total} & & & & \textbf{3{,}787} & \textbf{4{,}472} & \textbf{4{,}536} & \textbf{4{,}388} & \textbf{5{,}963} & \textbf{23{,}146} \\
\bottomrule
\end{tabular}
\caption{The \NTASKS\ tasks of \BENCHMARK: evaluation metric, human audit, and size.
Single- and multi-page variants of a task share a row, a metric and a score. \textbf{A\,+\,B} is the percentage of reference examples a human annotator rated correct (\textsc{a}) or acceptable (\textsc{b}) in the audit of \S\ref{ssec:dataset_creation}, out of 250 judgements per task; it measures the quality of the \emph{data}, not of any model. $^{\dagger}$Translation was audited by translation direction rather than by language, so its judgements are not on the same 250-per-task basis and we do not report a comparable figure. %
Appendix~\ref{app:generation} describes how each task is generated.}
\label{tab:tasks}
\end{table*}

\subsection{Statistics and Analysis}
The benchmark comprises \DATASETSIZE examples spanning five languages and \NTASKS task types, each generated from a distinct source document. Table~\ref{tab:tasks} reports the full task-by-language breakdown alongside each task's metric and audit score.

\paragraph{Language distribution.}
Coverage is broadly balanced but skewed towards Italian (5{,}963, $25.8\%$); German is least represented (3{,}787, $16.4\%$). The skew stems chiefly from three tasks (Named-entities Extraction, QA, Summarization), whose Italian partitions are two to three times larger than the other languages'.

\paragraph{Task distribution.}
Grouped into seven categories, examples concentrate in QA (6{,}205), Document Parsing (4{,}509) and Information Extraction (3{,}468), with Classification the smallest (1{,}000). At the task level, Named-entities Extraction (2968), QA (2700), Summarization (2574), and Translation (2380) dominate, while the three classification- and passage-oriented tasks are capped at 500 each.

\paragraph{Domain and document coverage.}
To understand documents distribution, we automatically tag examples with a rule-based domain classifier, which assigns one of 77 domains to 22{,}146 of the \DATASETSIZE examples ($95.7\%$), with a pronounced long tail (Figure~\ref{fig:domain_distribution}, Appendix~\ref{app:domain_classification}). Law (4{,}358), the economy (3{,}310), politics (1{,}889), medicine (1{,}854) and education (1{,}547) jointly account for $58.5\%$ of the tagged examples, while 49 domains contribute fewer than 100 examples each. Documents are predominantly single-page (mean 1.40 pages); the 3{,}196 multi-page documents average 3.93 pages.

\section{Evaluation Protocol}

\paragraph{Baselines.}
We benchmark \NMODELS\ open-weight VLMs spanning model families, architectures (dense and Mixture-of-Experts) and parameter scales from 1.7B to 235B total parameters. The suite spans the Qwen, InternVL, GLM and Gemma families across their released sizes, three OCR-specialised systems, and several general-purpose and multilingual-oriented models. Appendix~\ref{app:models} lists every evaluated system with its citation.

For all models we use the officially released instruction-tuned checkpoints and default chat templates, running inference with greedy decoding (temperature $0$, $\mathrm{top}_p = 1$, fixed seed). We run inference with vLLM in offline mode, in \texttt{bfloat16} precision, on NVIDIA GPUs, one node of eight GPUs per model. We set tensor parallelism to the smallest degree whose weights fit within a fixed fraction of per-GPU memory, and run the remaining GPUs as independent engine replicas over disjoint slices of the (task, language) grid. We derive each context window from the measured page resolution rather than assuming it: we cap images at $2.36$ Mpx and size the window for the largest number of pages in any record (five) plus the prompt and the generation budget. We cap output at $1024$ tokens for short-form tasks and $8192$ for full-page transcription, translation and extraction.

\subsection{Evaluation Metrics}
\label{ssec:eval_metrics}

A single metric across a task suite this heterogeneous would be misleading: exact match punishes the discursive answers our references contain, while character-level distance is meaningless for a summary. We therefore assign each task the metric its answer space admits, and report two guards alongside every score, whether the model produced a parseable answer at all, and whether the response was truncated, so that a low score can be read as a failure of capability rather than of format. We also apply a minimum-detectable-difference criterion per cell: where the number of records cannot separate two models at the resolution we report, we mark the cell underpowered. Two classification tasks fall out this way; we report them per task but leave them out of the per-language averages, where each cell is five times smaller.

Each task has a dedicated metric. For transcription we use $\mathrm{CER}_{\mathrm{sim}} = 1 - \min(1, \mathrm{Lev}(\hat{y},y)/|y|)$, normalising Markdown entries. Translation uses chrF++ and summarization chrF, which measures surface overlap against a single reference. Conditional Summarization uses a constraint satisfaction rate, since its constraints are verifiable without a reference. QA uses token recall, as MDU references are discursive and exact match penalizes correct but verbose answers. NE Extraction uses micro-$F_1$ over the value set a response reports. For Layout Extraction we score the page words a predicted box covers rather than the geometry of the box itself (\S\ref{ssec:results_and_discussion}), rescaling each response by its largest coordinate first, since most models emit the $0$--$1000$ grid they were trained on rather than the page fractions the references use. Because no answer format is guaranteed, we report coordinate coverage, the share of responses from which any box could be parsed, as a guard. We normalize all predictions before scoring: NFC composition, whitespace collapsing, and removal of delimited reasoning blocks.
\paragraph{Reproducibility.} We fix the seed at $0$ and decode greedily. Every model runs in \texttt{bfloat16} with prefix caching, eager execution, at most $32$ concurrent sequences and $0.85$ of per-GPU memory; we resample images with a Lanczos filter to the $2.36$ Mpx cap and score the final turn. A cell is reportable when its minimum detectable difference at $80\%$ power stays below $0.05$, underpowered up to $0.10$, and not reportable above that or below $40$ records. Interval estimates are BCa bootstrap over $10{,}000$ resamples clustered by source document. We release the evaluation set, the per-item scores and the harness.

\subsection{Results and Discussion}

\label{ssec:results_and_discussion}

\begin{table*}[t]
\centering\scriptsize\setlength{\tabcolsep}{3.5pt}
\rowcolors{2}{}{black!5}
\begin{tabular}{@{}lcccccccccccccc@{}}
\toprule
\multirow{2}{*}{\textbf{Model (params)}} & \multirow{2}{*}{\textbf{AVG}} & \multicolumn{2}{c}{\textbf{Parsing}} & \multirow{2}{*}{\textbf{Layout}} & \multicolumn{2}{c}{\textbf{Inf.\ Extr.}} & \multicolumn{2}{c}{\textbf{Classif.}} & \multicolumn{3}{c}{\textbf{Question Answering}} & \multicolumn{2}{c}{\textbf{Summarization}} & \multirow{2}{*}{\textbf{Trans}} \\
\cmidrule(lr){3-4}\cmidrule(lr){6-7}\cmidrule(lr){8-9}\cmidrule(lr){10-12}\cmidrule(lr){13-14}
 &  & \textbf{Text} & \textbf{MD} &  & \textbf{NER} & \textbf{Pass} & \textbf{Type} & \textbf{Dom} & \textbf{QA} & \textbf{TabQA} & \textbf{FigQA} & \textbf{Sum} & \textbf{CSum} &  \\
\midrule
\rowcolor{white}\multicolumn{15}{@{}l}{\textbf{\textit{Small ($<$10B)}}} \\[1pt]
EuroVLM (9B) & 40.6 & 24.6 & 41.5 & 11.6 & 20.5 & 39.5 & 51.1 & 58.7 & 53.9 & 44.8 & 42.5 & 40.2 & \textbf{77.0} & 21.4 \\
GLM-4.1V (9B) & 46.1 & 56.9 & 50.8 & 37.2 & \underline{32.6} & 45.7 & 52.5 & 46.3 & 36.7 & 53.4 & 44.6 & 42.8 & 59.4 & 39.9 \\
Granite-Vision-3.3 (2B) & 35.4 & 52.9 & 10.6 & \underline{41.9} & 19.1 & 35.9 & 23.5 & 39.7 & 51.1 & 31.5 & 38.0 & 24.9 & 68.1 & 23.4 \\
Idefics3 (8B) & 35.5 & 31.6 & 41.2 & 8.2 & 19.4 & 32.2 & 42.8 & 54.1 & 47.9 & 32.3 & 34.3 & 29.2 & \underline{71.6} & 17.2 \\
Keye-VL (8B) & 50.9 & 53.6 & 50.5 & \textbf{42.0} & 31.2 & \underline{46.0} & 60.7 & 67.3 & \underline{54.6} & \underline{58.6} & \underline{51.7} & \underline{48.8} & 59.4 & 37.7 \\
MiMo-VL (7B) & \underline{52.7} & 65.2 & 76.4 & 28.0 & \textbf{36.2} & 39.2 & \textbf{71.6} & \textbf{75.8} & 38.8 & 57.4 & 44.1 & 42.1 & 64.0 & \textbf{46.6} \\
Nanonets-OCR-s (3B) & 49.9 & \textbf{81.8} & \textbf{81.5} & 22.1 & 19.7 & 43.0 & 50.1 & 61.0 & 53.8 & 53.9 & 46.3 & 45.4 & 58.0 & 32.8 \\
Ovis2.5 (9B) & 47.4 & 49.2 & 51.4 & 30.3 & 29.6 & 40.3 & 60.6 & 71.1 & 35.9 & 54.9 & 47.6 & 40.7 & 65.0 & \underline{40.0} \\
Qwen2-VL (2B) & 35.3 & 43.9 & 31.1 & 5.7 & 11.4 & 39.0 & 28.3 & 40.1 & 50.3 & 44.9 & 40.0 & 35.5 & 64.9 & 24.1 \\
dots.ocr (1.7B) & 35.0 & 75.6 & 73.6 & 6.5 & 0.3 & 43.7 & 0.9 & 0.7 & 39.6 & 57.5 & 35.0 & 45.0 & 51.3 & 25.2 \\
olmOCR-2 (7B) & \textbf{56.1} & \underline{81.5} & \underline{79.9} & 24.9 & 24.2 & \textbf{49.5} & \underline{64.5} & \underline{72.3} & \textbf{59.8} & \textbf{63.8} & \textbf{53.4} & \textbf{51.2} & 65.2 & 38.3 \\
\addlinespace[2pt]
\rowcolor{white}\multicolumn{15}{@{}l}{\textbf{\textit{Medium (10--40B)}}} \\[1pt]
Gemma3 (27B) & 57.6 & 63.6 & 44.7 & 40.1 & 43.1 & \underline{46.7} & 70.0 & 76.0 & 55.3 & 61.8 & 54.0 & 48.8 & \textbf{90.3} & 53.9 \\
Gemma4 (31B) & 65.4 & 73.2 & 70.3 & 52.7 & \underline{52.4} & 43.6 & \underline{83.7} & \underline{83.4} & 59.4 & 63.4 & 59.7 & \textbf{53.2} & \underline{89.4} & 66.2 \\
InternVL3.5 (38B) & 62.1 & 76.5 & 78.8 & 54.0 & 42.0 & 43.8 & 72.3 & 75.6 & 60.6 & 63.2 & 53.2 & 50.1 & 87.5 & 50.0 \\
Kimi-VL (16B-A3B) & 46.8 & 60.4 & 69.4 & 5.0 & 22.2 & 35.9 & 60.3 & 61.5 & 52.0 & 42.9 & 42.5 & 45.0 & 75.5 & 36.7 \\
Pixtral (12B) & 51.4 & 67.1 & 71.6 & 28.4 & 28.0 & 36.4 & 54.3 & 57.1 & 54.4 & 55.5 & 45.6 & 48.0 & 77.2 & 44.3 \\
Qwen3.5 (27B) & \underline{68.1} & \underline{76.8} & \underline{84.9} & \textbf{61.6} & \textbf{53.5} & 46.5 & 80.8 & \textbf{84.1} & \underline{62.2} & \underline{68.8} & \underline{63.9} & \underline{53.1} & 82.2 & \underline{67.1} \\
Qwen3.6 (27B) & \textbf{69.0} & \textbf{80.6} & \textbf{88.5} & \underline{61.1} & 52.3 & \textbf{48.4} & \textbf{83.8} & 82.4 & \textbf{64.4} & \textbf{70.7} & \textbf{64.9} & 52.9 & 78.6 & \textbf{68.3} \\
\addlinespace[2pt]
\rowcolor{white}\multicolumn{15}{@{}l}{\textbf{\textit{Large ($>$40B)}}} \\[1pt]
GLM-4.5V (106B-A12B) & 59.2 & \underline{76.8} & 77.6 & \underline{44.0} & 41.8 & 48.4 & 60.1 & 67.1 & 49.5 & 63.2 & 55.2 & \underline{53.6} & 68.5 & \underline{63.4} \\
InternVL3 (78B) & 55.8 & 54.3 & 53.9 & 29.4 & 41.0 & 40.2 & 75.4 & 76.8 & 56.6 & 61.3 & 54.0 & 47.8 & \underline{87.9} & 47.0 \\
Qwen2.5-VL (72B) & \textbf{66.8} & \textbf{82.2} & \textbf{81.9} & \textbf{52.1} & \textbf{48.4} & \underline{48.7} & \underline{75.7} & \underline{79.6} & \underline{63.0} & \underline{67.3} & \underline{59.2} & \textbf{56.1} & \textbf{91.4} & 63.4 \\
Qwen3-VL (235B-A22B) & \underline{66.3} & 76.8 & \underline{80.0} & 38.6 & \underline{47.7} & \textbf{52.6} & \textbf{75.9} & \textbf{81.5} & \textbf{68.1} & \textbf{76.3} & \textbf{65.4} & 53.1 & 78.4 & \textbf{67.6} \\
\midrule
Gemini 3.8 Flash (commercial, API) & 65.8 & 73.8 & 71.1 & 44.5 & 50.9 & 46.0 & 79.5 & 79.0 & 59.3 & 74.1 & 65.2 & 53.3 & 87.0 & 71.7 \\
\bottomrule
\end{tabular}
\caption{Per-task scores on \BENCHMARK. To keep the table readable we show \textbf{one model per family}, the best-scoring one; the complete table for all \NMODELS\ open-weight systems is in Table~\ref{tab:results-full}. Models are grouped by parameter scale and ordered alphabetically by family within each group. AVG is the unweighted mean over the \NEVALTASKS tasks. Best per column \emph{within each size group} in bold, runner-up underlined; the commercial system is a single reference point and is not part of that comparison. Single- and multi-page variants are merged. All scores $\times100$.}
\label{tab:results}
\end{table*}

\begin{figure}[t]
\centering
\includegraphics[width=\columnwidth]{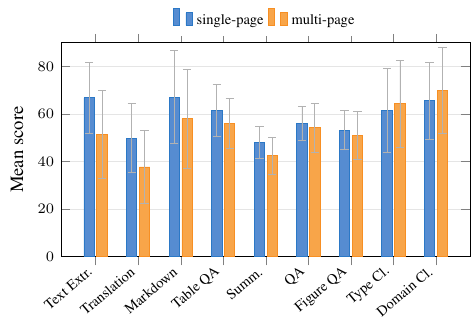}
\caption{Single-page versus multi-page performance, averaged over the \NMODELS\ evaluated models. Bars are means, whiskers one standard deviation across models. Tasks are ordered by the size of the drop. The nine tasks with paired variants are shown; Layout, NE and Passage Extraction and Conditional Summarization exist in single-page form only. The two classification tasks are the only ones where the multi-page variant is easier.}
\label{fig:singlemulti}
\end{figure}

Table~\ref{tab:results} reports per-task scores. Overall, only eight of the \NMODELS\ models exceed 65 on the \BENCHMARK\ average and none reaches 70, so the suite is neither saturated nor out of reach.  \textit{Qwen3.6 (27B)} leads with an average of 69.0, ahead of \textit{Qwen3.5 (27B)} (68.1); \textit{Qwen2.5-VL (72B)} follows at 66.8 and \textit{Qwen3-VL (235B-A22B)} at 66.3. The two best systems in the suite are both 27B models, and each beats every model above 40B: scale buys a general advantage, but it does not order the leaderboard. Parameter count alone explains $R^2=0.41$ of the variance in the average, model family $0.85$. We also ran one commercial system under the same protocol: \textit{Gemini 3.8 Flash} averages 65.8, seventh overall, so the leaderboard is not behind the one proprietary reference we could measure. Its profile differs from the open-weight leaders rather than simply trailing them: it ranks first in the suite on Translation and within the top five on Summarization, Table QA and Figure QA, but twenty-seventh on Markdown Extraction and nineteenth on Text Extraction, the two tasks where transcription fidelity decides the score.  

\paragraph{Capabilities do not come together.}
Strong performance on one task family does not carry to others, and the table shows it directly. \textit{Nanonets-OCR-s (3B)} reaches $81.8$ on Text Extraction, second in the suite, yet scores $19.7$ on entity extraction and $32.8$ on translation and averages $47.9$: thirty-three models outrank it overall, the weakest of them $28$ points behind it on transcription. \textit{dots.ocr (1.7B)} is the sharper case: $75.6$ on Text Extraction and $73.6$ on Markdown against $0.3$ on entity extraction, because it answers every prompt with its own layout-dump format regardless of the instruction. Parsing ability and instruction following are separable, and a benchmark that reports only an average hides which one a model has. The pattern holds across the whole suite: controlling for general model quality, the mean partial correlation between tasks falls from $+0.66$ to $+0.09$ (Appendix~\ref{app:task_independence}). Headroom is unevenly distributed too. Layout Extraction and entity extraction leave the most of any task, averaging $34.6$ and $34.8$ across systems against $77.4$ on Conditional Summarization.

\paragraph{The generator family.}
\BENCHMARK's instruction--response pairs are produced by a Qwen model (\S\ref{ssec:dataset_creation}), and Qwen models occupy nine of the ten top positions in Table~\ref{tab:results-full}. The concern this raises is self-preference: that the benchmark rewards resemblance to its generator rather than document understanding. The generator is not at the top: evaluated under the same protocol, \textit{Qwen3.5 (122B-A10B)} averages $67.4$, behind both 27B models, where a benchmark that merely rewarded resemblance to it would rank it first. We measure the residual anyway. Regressing the average on $\log_{10}$ parameters and an indicator for the Qwen family leaves a residual advantage of $+7.8$ points at equal scale, larger than the mean gap of $5.3$ between a model's best and worst language.

Its distribution, however, is not what a genuine capability gap would look like. A real advantage would be spread across tasks; a stylistic one concentrates where the metric rewards surface resemblance to the reference. The advantage is $+16.0$ on Markdown Extraction and between $+4.0$ and $+10.7$ on every other reference-scored task. On the two tasks whose metrics do not compare surface form against a reference (Layout Extraction, scored by the page words a box covers, and Conditional Summarization, scored by constraint satisfaction) it is $+4.9$ and $+0.5$. Averaged, that is $+8.7$ points on reference-scored tasks against $+2.7$ on format-independent ones, a factor of three.

\paragraph{Scoring layout on text rather than geometry.}
Layout Extraction asks a model to localise named elements of a page, and the obvious score, $F_1$ at an IoU threshold over matched boxes, measures the wrong thing: a third of the reference boxes are a single line of text, under $0.02$ in height, where an error of one line drives IoU to zero. We therefore score a prediction on the page words its boxes cover rather than on their geometry, using the word positions in the corpus. Scores move from a 
  $0$--$35$ range to $0$--$61$, and the ordering changes where it should: a model answering with one bare box where three were requested drops four places once its geometric credit stops counting. Appendix~\ref{app:layout_components} reports both scores with two diagnostics.

We therefore read part of the Qwen ordering in Table~\ref{tab:results} as agreement with the generator's output conventions rather than document understanding alone, and offer the two format-independent tasks as a conservative cross-family comparison. The effect belongs to synthetic benchmark construction in general rather than to this dataset in particular, and quantifying it is what lets a reader discount it.

\paragraph{Difficulty varies by domain, and document properties do not explain it.}
\BENCHMARK's documents span 77 domains, and scores differ across them: pooling every model and taking the macro mean over tasks, the gap between the easiest and the hardest domain is $7.5$ points. Part of that could be composition rather than domain, since domains do not contain the same mix of tasks or languages, so we regress out task and language and rank domains on the residuals. The gap narrows only to $5.9$ points, and the two rankings agree closely ($\rho=0.85$ over the 28 domains with at least $5{,}000$ scored items): $79\%$ of the raw spread belongs to the domain itself.

What the spread is \emph{not} explained by is more informative than the spread. We correlated per-domain difficulty against every document property we can measure, and none accounts for it: the strongest, the number of layout elements, reaches only $\rho=0.36$, and prompt and reference length stay below $0.11$. The signs also run against intuition: longer, denser documents score slightly \emph{higher}. Most telling, neither the difficulty the generator assigned to each example nor the share it labelled \emph{hard} correlates with the difficulty models experience ($\rho=-0.22$ and $-0.20$). A synthetic pipeline can control what it labels difficult; it does not thereby control what is difficult.

Qualitatively, the residual ordering tracks genre rather than size. The hardest buckets are food ($-0.032$), armed forces ($-0.031$) and culture ($-0.030$); the easiest are ecology ($+0.027$), entertainment ($+0.022$) and education ($+0.015$). This reads as a layout and register effect: the easy end is prose-heavy writing with conventional page structure, the hard end documents built out of menus, forms, posters and tables, whose surface departs from running prose. The two classification tasks, whose own labels are domains, are excluded (Appendix~\ref{app:domain_classification}). 

Family effects are consistent: Qwen3.5 outperforms Qwen3-VL, and the advantage concentrates in Type Classification, Domain Classification, Markdown Extraction and NER: tasks that require structured, short, language-faithful output rather than open-ended generation.

\paragraph{English leads, German trails, and scale does not close the gap.}
Per-language averages for every evaluated system are in Table~\ref{tab:languages-full} (Appendix~\ref{apx:full_results}). English is the strongest language for 34 of \NMODELS\ models and German the weakest for 42: on average 55.7 for English, 53.8 for Spanish, 53.3 for Italian, 52.5 for French and 50.6 for German. The gap does not close with scale: mean spread is 5.7 points below 10B, 4.8 between 10B and 40B and 5.2 above. Larger models raise all five languages together rather than closing the distance between them.

A natural objection is that this gap reflects \emph{what} each language is about rather than the language itself: the domain composition of \BENCHMARK\ differs sharply across languages, with law accounting for $4.9\%$ of English examples and $35.9\%$ of Italian ones. It does not. Standardising each language to a common domain composition, at a fixed task mix, moves the mean best--worst spread by less than $0.001$ and leaves the English advantage over the remaining four languages unchanged. 

\paragraph{Multi-page evidence is expensive.}
Nine of the \NEVALTASKS\ tasks have paired single- and multi-page variants, which isolates a cost benchmarks built from single pages cannot see: what a model loses when the evidence is spread across a page sequence rather than one image (Figure~\ref{fig:singlemulti}). The cost is real and large. Averaged over models, Text Extraction drops 15.3 points from single to multi-page, Translation 12.1 and Markdown Extraction 9.1; Table QA drops 5.4 and Summarization 5.8. The effect is near-universal: 41 of \NMODELS\ models decline on Text Extraction, and on Translation all \NMODELS\ do. The two classification tasks are the only ones that move the other way, which is consistent with additional pages supplying redundant evidence for a single label while multiplying the content that transcription and translation must carry faithfully.

\section{Conclusions}
\label{sec:conclusion}
{\az We introduced \BENCHMARK, a multilingual Document Understanding benchmark of \DATASETSIZE examples over \NTASKS tasks and five languages, and used it to evaluate \NMODELS open-weight VLMs from 1.7B to 235B parameters.

Three findings stand out. The benchmark is far from saturated: the best average is 69.0, only eight of the \NMODELS\ evaluated systems reach 65, and layout and entity extraction defeat nearly every system. Scale helps but does not determine performance: parameter count alone explains $R^2=0.41$ of the variance in the average and model family $0.85$. The two best models overall are 27B and each outperforms every model above 40B. Capabilities do not improve jointly: strength on one family, such as transcription for OCR-specialized models, does not transfer, and multi-page inputs degrade every task but classification. English leads on average and German trails for 42 of \NMODELS\ models, so English-only evaluation overstates usability.

We release \BENCHMARK, its per-item scores and the evaluation harness, and hope it drives progress where current VLMs remain weakest: layout reasoning, entity extraction and multi-page understanding.
}

\section*{Limitations}

Our \emph{linguistic scope} is five European languages in the Latin script, all comparatively high-resource, so our cross-lingual findings may not carry to other scripts, morphologies or reading orders, nor to genuinely low-resource settings.

\emph{Document availability and licensing} shape the corpus. We draw on crawlable FinePDFs filtered to permissive licenses, which skews the domain distribution towards law, economy and politics and leaves some tasks and languages dependent on documents of unspecified license terms.

Our leaderboard is \emph{almost entirely open-weight}. The single commercial model is a reference point, not a survey of proprietary systems, and it is not comparable in every respect: its parameter count is undisclosed, we reached it through an API at a pinned provider tier rather than running it under our own decoding settings, and it ran with a low reasoning effort that has no counterpart in the open-weight runs. The provider also
flagged $2.1\%$ of its generations as incomplete while returning usable text, which we kept and scored.

\emph{The generation budget on Layout Extraction} caps responses at $1{,}024$ tokens, which $18\%$ reach. The cap does not bind a correct answer, since $99\%$ of references fit within it; truncation marks verbosity, and verbose responses score lower because the
metric normalises by the larger of the predicted and reference box counts. For those models the score should be read with the coordinate coverage we report beside it.

Finally, we rely on \emph{automatic metrics} with greedy decoding and fixed prompts. Surface-overlap measures may under- or over-penalize semantically valid variation, and a single prompt per task does not probe sensitivity to instruction phrasing.

\section*{Ethical Considerations}
\paragraph{Document provenance and licensing.}
\BENCHMARK\ is built over PDFs from FinePDFs, a crawl of the public web. We resolve the
licensing status of every source document, extending the search to the hosting site's
terms-of-use and legal-notice pages when the document carries no statement of its own, and we
exclude every document whose licence is non-permissive or whose site's \texttt{robots.txt}
restricts crawling or AI-related use (\S\ref{ssec:dataset_creation}). Explicitly permissive
licences are scarce: they cover $2.8\%$ of the crawl. To keep the under-represented tasks and
languages usable we also drew on documents whose site permits crawling but which carry no
explicit licence statement, and that subset is the part of the corpus whose redistribution we
cannot ground in an explicit licence grant. We therefore release it by reference, as source URL and
document hash together with the generated instruction and reference, and redistribute document
images only for the explicitly permissively licensed portion.

\paragraph{Personal data.}
The source documents are real and public, and documents of this kind name people: authors of
reports, parties to legal decisions, officials. We neither sought out nor removed such
mentions, and the generated instructions and references can therefore repeat a name that the
page already carries. We publish no data that is not already public at the source URL, and the
reference-based release above means a document withdrawn at source does not persist through
us.

\paragraph{Human annotation.}
The audit of \S\ref{ssec:dataset_creation} and the pre-generation validation were carried out
by native or near-native speakers of the five languages compensated as part of their employment, who rated references against the page
images.

\paragraph{Intended use.}
\BENCHMARK\ is an evaluation resource, not training data, and its scores measure what a model
produces under one fixed prompt per task with greedy decoding. They are not a statement about
deployment readiness: a high average here does not establish that a system is safe to run over
documents that carry legal or medical consequences for the people named in them. The references
are synthetic and come from a single generator family, an effect we measure rather than assume
(\S\ref{ssec:results_and_discussion}); readers comparing systems from that family against others
should discount the leaderboard by the margin we report there.

\bibliography{custom}

\ifwithappendix
\appendix

\section{Generation Pipeline per Task}
\label{app:generation}

\S\ref{ssec:tasks} states what each task asks of a model. This appendix records how its examples were produced, since the construction is what a reader needs in order to judge what the references can and cannot support. All tasks draw their instructions and responses from the generator described in \S\ref{ssec:dataset_creation}; what differs is the number of generation steps and the deterministic filtering applied before them.

\paragraph{Single-pass tasks.} Summarization, Conditional Summarization, Translation and
Type Classification are produced in one pass: a task-specific seeding prompt, usually translated across the five languages, is paired with the page image. For Conditional Summarization the sampled constraint is tied to a quantitative measurement (number of words or characters for free text, items for lists, rows for Markdown tables, concepts for key-concept lists) so that satisfaction can be checked mechanically afterwards.

\paragraph{Parametric sampling.} Table QA, Figure QA, Passage Extraction and Layout
Extraction sample a parameter vector per example before generation, which is what produces within-task variety rather than many rephrasings of one prompt. The parameters cover difficulty (from explicitly stated facts to inferential content), instruction mode (question or command), question type, answer style, and language; Passage Extraction adds an optional persona that modulates the voice of the query, following \citet{ge2406scaling}. Each of these tasks is preceded by a deterministic filter: a page must contain a table for Table QA, a figure above a minimum size for Figure QA (logos and icons are excluded) and a minimum character count for Passage Extraction. Layout Extraction is seeded differently: a YOLO-based detector first extracts every textual and non-textual element with its normalised box, and the generator receives that layout alongside the page, so the reference is grounded in a detector's output rather than in the generator's own reading of the page. Responses are then validated structurally, and examples whose reported elements do not match the extracted layout are discarded.

\paragraph{Multi-step tasks.} NE Extraction required a staged pipeline
rather than a single pass, because a single end-to-end step produced inconsistent annotations. NE Extraction runs in three steps: the generator first identifies entity mentions on the page guided only by broad category indications; it then assigns a category to each mention and structures the output against a fixed JSON schema; finally it writes a natural-language instruction consistent with the entities actually extracted, so that the request is reverse-engineered from the annotation instead of the annotation being produced from a fixed prompt. Decoupling identification, categorisation and instruction writing is what keeps the three from contaminating each other.

\paragraph{Validation and discards.} Parsing tasks are validated by fence detection:
responses without the expected fences, or containing only the sentinel for untranscribable pages, are recorded as extraction failures and dropped. Tasks with a structured response are parsed against their expected format and dropped on failure. Retained examples are stored in chat format, pairing page image and instruction as the user turn with the response as the assistant turn.

\section{Conditional Summarization Constraints}\label{app:condsum}
Table~\ref{tab:summary_types} lists the fourteen constraints a Conditional Summarization instruction can carry. We sample one per example, and score the response against it directly, without a reference.

\begin{table}[htbp]
\centering\small
\begin{tabular}{@{}l p{4.9cm}@{}}
\hline
\textbf{Type} & \textbf{Description} \\
\hline
single\_sentence & summarize in a single short sentence \\
bulleted\_list & summarize as a bulleted list \\
numbered\_list & summarize as a numbered list \\
actions\_intentions & summarize with a focus on actions and intentions \\
key\_info & summarize focusing only on key information \\
n\_key\_concepts & summarize in n key concepts \\
entities & summarize with a particular focus on the entities involved \\
markdown\_table & summarize as a markdown table \\
conversational & summarize adopting a conversational style \\
telegraphic & summarize adopting a telegraphic style \\
journalistic & summarize adopting a journalistic style \\
administrative & summarize adopting an administrative/bureaucratic style \\
plain\_language & summarize adopting plain language, as if intended for a foreign learner of the language \\
casual & summarize adopting a casual style, as if written in a chat \\
\hline
\end{tabular}
\caption{Summary types and their descriptions.}
\label{tab:summary_types}
\end{table}

\section{One Example per Task}
\label{app:task_examples}
Table~\ref{tab:task-examples} gives, for each task, the instruction a model receives and the reference it is scored against, both shortened. The page image that accompanies every instruction is omitted. Examples are drawn from different languages to show that instructions are written in the language of the document.

\begin{table*}[htbp]
\centering\scriptsize
\setlength{\tabcolsep}{4pt}\renewcommand{\arraystretch}{1.15}
\begin{tabular}{@{}llp{0.385\linewidth}p{0.385\linewidth}@{}}
\toprule
\textbf{Task} & \textbf{Lang} & \textbf{Instruction} & \textbf{Reference} \\
\midrule
Text Extraction & \texttt{fr} & Peux-tu lire et recopier tout le texte ? & digilor FICHE TECHNIQUE BORNE IPAD COLUMN Cette borne iPad COLUMN au design tout en courbe s'intègre parfaitement sur vos points de vente ou lors de vos salons. Elle accueille votre [...] \\
\addlinespace[2pt]
Markdown Extraction & \texttt{fr} & Convertissez cette page de document en Markdown. & Et après en avoir délibéré, > CONSTATE effectivement un allongement de la durée du chantier qui devrait s'achever en juillet 2024. > DÉCIDE la signature d'un marché complémentaire avec le [...] \\
\addlinespace[2pt]
Layout Extraction & \texttt{it} & Individua le didascalie presenti in questa pagina e forniscimi le loro posizioni. & In questa pagina è presente una didascalia con il testo "Sala del Rifrattore". La sua posizione corrisponde alla regione [x1=0.406, y1=0.316, x2=0.593, y2=0.333] in coordinate normalizzate, [...] \\
\addlinespace[2pt]
\addlinespace[2pt]
NE Extraction & \texttt{it} & Analizza l'immagine allegata per identificare le entità presenti. Restituisci il risultato come un oggetto JSON in cui ogni chiave è il testo esatto dell'entità trovata e ogni valore è la [...] & \{ "Repubblica Italiana": "country", "COMUNE DI CASTELPIZZUTO": "city", "Provincia di Isernia": "state", "n. 2 del 23-02-2016": "DATE", "S2": "ID", "Folgara": "LOCATION" \} \\
\addlinespace[2pt]
Passage Extraction & \texttt{en} & What specific modification does the guide recommend when ending the Butterfly stitch to avoid creating an unstable loop? & After numerous additional stitches, I'm ready to end the Butterfly stitch, so I tighten up this last loop, but this time, instead of making a loop to stick through this last loop, I want to [...] \\
\addlinespace[2pt]
Type Classification & \texttt{en} & Operate as a document classifier. You receive ONE page of a document as an image. TASK: assign ONE label - the document TYPE (structure/nature). Pick exactly one of the 10 type classes [...] & justification,has\_main\_body\_text,type\_confidence,type "The document is a formal letter addressed to Dr Brent Layton (Chair of the Electricity Authority) from a group of stakeholders. It [...] \\
\addlinespace[2pt]
Domain Classification & \texttt{en} & Act as a document classifier. A SINGLE page of a document is provided to you as an image. TASK: assign ONE label - the document DOMAIN (subject/topic). Pick exactly one of the 10 domain [...] & \{ "domain": "education", "domain\_confidence": "high", "has\_main\_body\_text": true, "justification": "The document is a schedule for 'Summer Reading 2023' programs at various public [...] \\
\addlinespace[2pt]
QA & \texttt{it} & Chi ha presieduto la seduta del Nucleo di Valutazione tenutasi il 8 febbraio 2021? & La riunione è stata guidata dal Coordinatore Stephen John Taylor. \\
\addlinespace[2pt]
Table QA & \texttt{es} & Identifica los nombres de los candidatos que se encuentran inmediatamente encima y debajo de 'José Miñano Sánchez' en la tabla de resultados. & En la lista de resultados, el candidato que aparece inmediatamente encima de 'José Miñano Sánchez' es 'Ester José Marín Lidón'. El candidato que se encuentra inmediatamente debajo es [...] \\
\addlinespace[2pt]
Figure QA & \texttt{it} & Qual è la differenza di velocità di connessione internet prevista tra la tecnologia attuale e quella in fibra ottica, come specificato nel testo accanto all'immagine panoramica di Cortina? & L'articolo indica che la velocità attuale è di 7 megabit al secondo. Con l'arrivo della fibra ottica, la velocità salirà a 100 megabit al secondo, con una previsione futura di raggiungere i [...] \\
\addlinespace[2pt]
Summarization & \texttt{it} & Estrai il succo da questo testo. & Il documento presenta il Bilancio di Previsione 2017 del Comune di Casaleggio Novara relativo al Servizio 01 - Peso Pubblico. I ricavi, derivanti dai diritti di peso e misura pubblica [...] \\
\addlinespace[2pt]
Conditional Summ. & \texttt{en} & Summarize the main actions and intentions depicted in the attached image, ensuring the response is at most 50 words. & The Harvest of the Month program provides registered schools with a program manual, fun and nutrition facts, promotional language for school and community communications, curricular [...] \\
\addlinespace[2pt]
Translation & \texttt{it} & Puoi tradurre il testo di questa pagina in inglese? & - December 8th, Immaculate Conception; - December 25th, Christmas Day; - December 26th, St. Stephen's Day; - January 1st, New Year's Day; - January 6th, Epiphany; - Easter Monday; - April [...] \\
\addlinespace[2pt]
\bottomrule
\end{tabular}
\caption{One example per task of \BENCHMARK. Instructions and references are shortened; \texttt{[...]} marks a cut. The classification tasks enumerate their admissible labels in the instruction, which is why theirs are the longest.}
\label{tab:task-examples}
\end{table*}

\section{Layout Extraction: Reported Metric and Alternatives}
\label{app:layout_components}

Layout Extraction is scored on the page words a predicted box covers rather than on the geometry of the box. The reasons are properties of the task rather than of the models. The reference is generated by a detector and inherits its localisation error; a third of its boxes are below $0.02$ in height, so an $F_1$ at IoU $0.5$ has almost no tolerance for sub-line error; and roughly half of a page is margin, which an area-based overlap penalises a model for including even though no text is at stake. Measured on words the same predictions score between $0$ and $61$ rather than between $0$ and $35$

Two diagnostics in Table~\ref{tab:layout_components} explain what remains. On the $1{,}411$ items ($61\%$) whose reference names the elements it localises, predictions can be matched to it by element text rather than by overlap, which separates identifying the requested element from localising it. Referent $F_1$ is low throughout: in most cases model and reference are describing different objects, and the task's phrasing (``the main title'' on a page whose figure carries its own caption) is genuinely ambiguous as to both which element is meant and at what granularity. The ratio of predicted to reference boxes shows the second failure mode: models over- and under-produce boxes in opposite directions, and the geometric metric rewarded the models that said least.

Two limits bound this. The word-level analysis needs the reference to name its elements, which it does for $61\%$ of items; beyond that the ambiguity cannot be resolved by rescoring, because it is the reference that is under-specified. And the cause upstream is the answer format: free-form prose with coordinates embedded obliges neither side to state what is being localised. A structured answer, giving element type, text and box, would remove the ambiguity, the count miscalibration and the verbosity together, at the cost of regenerating the task.

\begin{table}[htbp]
\centering\footnotesize\setlength{\tabcolsep}{2.5pt}
\rowcolors{2}{}{black!5}
\resizebox{\columnwidth}{!}{%
\begin{tabular}{@{}lcccccc@{}}
\toprule
\textbf{Model (params)} & \textbf{$F_1$} & \textbf{P} & \textbf{R} & \textbf{$F_1@50$} & \textbf{Ref.\ $F_1$} & \textbf{$n_p/n_g$} \\
\midrule
EuroVLM (1.7B) & 0.002 & 0.00 & 0.00 & 0.000 & 0.028 & 0.36 \\
EuroVLM (9B) & 0.126 & 0.14 & 0.18 & 0.013 & 0.091 & 0.41 \\
GLM-4.1V (9B) & 0.367 & 0.47 & 0.37 & 0.101 & 0.186 & 1.53 \\
Gemma3 (4B) & 0.260 & 0.34 & 0.30 & 0.019 & 0.159 & 1.38 \\
Gemma4 (E2B) & 0.275 & 0.42 & 0.26 & 0.061 & 0.075 & 1.74 \\
Gemma4 (E4B) & 0.403 & 0.53 & 0.39 & 0.122 & 0.142 & 1.24 \\
Granite-Vision-3.3 (2B) & 0.431 & 0.56 & 0.45 & 0.256 & 0.001 & 0.08 \\
Idefics3 (8B) & 0.078 & 0.08 & 0.15 & 0.007 & 0.107 & 0.87 \\
InternVL3.5 (2B) & 0.242 & 0.34 & 0.26 & 0.032 & 0.140 & 1.02 \\
InternVL3.5 (4B) & 0.437 & 0.58 & 0.43 & 0.143 & 0.181 & 0.93 \\
InternVL3.5 (8B) & 0.431 & 0.57 & 0.42 & 0.112 & 0.225 & 1.01 \\
Keye-VL (8B) & 0.422 & 0.50 & 0.44 & 0.081 & 0.186 & 0.87 \\
MiMo-VL (7B) & 0.276 & 0.41 & 0.27 & 0.034 & 0.105 & 2.39 \\
Nanonets-OCR-s (3B) & 0.227 & 0.27 & 0.33 & 0.024 & 0.008 & 1.45 \\
Ovis2.5 (9B) & 0.308 & 0.39 & 0.33 & 0.023 & 0.174 & 1.55 \\
Qwen2-VL (2B) & 0.061 & 0.12 & 0.07 & 0.007 & 0.011 & 0.15 \\
Qwen2.5-VL (7B) & 0.273 & 0.38 & 0.27 & 0.061 & 0.094 & 1.23 \\
Qwen3-VL (2B) & 0.243 & 0.34 & 0.28 & 0.052 & 0.032 & 1.21 \\
Qwen3-VL (4B) & 0.374 & 0.46 & 0.40 & 0.073 & 0.140 & 1.49 \\
Qwen3-VL (8B) & 0.446 & 0.54 & 0.47 & 0.111 & 0.085 & 0.80 \\
Qwen3.5 (2B) & 0.237 & 0.32 & 0.29 & 0.035 & 0.063 & 2.30 \\
Qwen3.5 (4B) & 0.386 & 0.52 & 0.38 & 0.106 & 0.101 & 1.67 \\
Qwen3.5 (9B) & 0.415 & 0.54 & 0.41 & 0.136 & 0.055 & 2.32 \\
dots.ocr (1.7B) & 0.086 & 0.10 & 0.13 & 0.008 & 0.000 & 1.56 \\
olmOCR-2 (7B) & 0.256 & 0.33 & 0.27 & 0.058 & 0.039 & 0.69 \\
Gemma3 (12B) & 0.303 & 0.36 & 0.35 & 0.033 & 0.169 & 1.10 \\
Gemma3 (27B) & 0.396 & 0.48 & 0.43 & 0.038 & 0.183 & 1.46 \\
Gemma4 (26B-A4B) & 0.475 & 0.58 & 0.47 & 0.207 & 0.084 & 1.16 \\
Gemma4 (31B) & 0.520 & 0.61 & 0.51 & 0.234 & 0.097 & 1.26 \\
InternVL3.5 (14B) & 0.427 & 0.57 & 0.41 & 0.111 & 0.220 & 1.08 \\
InternVL3.5 (30B-A3B) & 0.512 & 0.65 & 0.50 & 0.218 & 0.200 & 0.93 \\
InternVL3.5 (38B) & 0.536 & 0.66 & 0.54 & 0.273 & 0.189 & 1.06 \\
Kimi-VL (16B-A3B) & 0.043 & 0.06 & 0.06 & 0.005 & 0.015 & 0.20 \\
Pixtral (12B) & 0.289 & 0.39 & 0.31 & 0.025 & 0.153 & 1.05 \\
Qwen2.5-VL (32B) & 0.482 & 0.56 & 0.50 & 0.146 & 0.245 & 1.55 \\
Qwen3-VL (30B-A3B) & 0.398 & 0.51 & 0.39 & 0.160 & 0.124 & 1.39 \\
Qwen3-VL (32B) & 0.371 & 0.46 & 0.38 & 0.072 & 0.248 & 1.34 \\
Qwen3.5 (27B) & 0.611 & 0.69 & 0.62 & 0.345 & 0.105 & 1.80 \\
Qwen3.5 (35B-A3B) & 0.418 & 0.52 & 0.41 & 0.174 & 0.084 & 1.70 \\
Qwen3.6 (27B) & 0.601 & 0.69 & 0.61 & 0.351 & 0.032 & 1.78 \\
Qwen3.6 (35B-A3B) & 0.463 & 0.58 & 0.46 & 0.196 & 0.114 & 1.69 \\
GLM-4.5V (106B-A12B) & 0.434 & 0.55 & 0.42 & 0.208 & 0.177 & 1.05 \\
InternVL3 (78B) & 0.291 & 0.42 & 0.27 & 0.085 & 0.159 & 0.86 \\
Qwen2.5-VL (72B) & 0.514 & 0.61 & 0.51 & 0.201 & 0.151 & 1.65 \\
Qwen3-VL (235B-A22B) & 0.382 & 0.50 & 0.36 & 0.133 & 0.125 & 1.32 \\
\bottomrule
\end{tabular}}

\caption{Layout Extraction, reported metric and its alternatives. \textbf{$F_1$} is the
harmonic mean of the precision (\textbf{P}) and recall (\textbf{R}) of the page words covered by the predicted boxes against those covered by the reference boxes, pooled over all items. Table~\ref{tab:results} reports the macro mean over the five languages instead, the convention of every other column there; the two differ by $0.5$ points at the median. \textbf{$F_1@50$} is the geometric alternative it replaced, $F_1$ at an IoU threshold of $0.5$ over a one-to-one box assignment. \textbf{Ref.\ $F_1$} and \textbf{$n_p/n_g$} come from a separate analysis on the $1{,}411$ items whose reference names the elements it localises: whether the requested elements were identified at all, and the ratio of predicted to reference boxes.}
\label{tab:layout_components}
\end{table}

\section{Evaluated Models}
\label{app:models}

We benchmark a broad set of open-weight VLMs spanning model families, architectures (dense and Mixture-of-Experts), and parameter scales ranging from 1.7B to 235B (total) parameters. We include the \textbf{Qwen} family: Qwen2-VL-2B \citep{Qwen2VL}, Qwen2.5-VL (7B, 32B, 72B) \citep{Qwen2.5-VL}, Qwen3-VL (2B, 4B, 8B, 32B, and the MoE variants 30B-A3B and 235B-A22B) \citep{qwen3technicalreport}, Qwen3.5 (2B, 4B, 9B, 27B, and the MoE 35B-A3B) \citep{qwen3.5}, and Qwen3.6 (27B and the MoE 35B-A3B). We further include the \textbf{InternVL} family, namely InternVL3-78B \citep{zhu2025internvl3exploringadvancedtraining} and InternVL3.5 (2B, 4B, 8B, 14B, 38B, and the MoE 30B-A3B) \citep{wang2025internvl3_5}, as well as \textbf{GLM-4.5V} (106B-A12B) and \textbf{GLM-4.1V-9B} \citep{vteam2026glm45vglm41vthinkingversatilemultimodal}, \textbf{Gemma~3} (4B, 12B, 27B) \citep{gemmateam2025gemma3technicalreport}, and \textbf{Gemma~4} (E2B, E4B, 31B, and the MoE 26B-A4B) \citep{gemmateam2026gemma4technicalreport}.

To assess models specialized for document understanding, we include three OCR-oriented systems: \textbf{olmOCR-2-7B} \citep{poznanski2025olmocr2unittest}, \textbf{Nanonets-OCR-s} (3B) \citep{Nanonets-OCR-S} and \textbf{dots.ocr} (1.7B) \cite{li2025dotsocrmultilingualdocumentlayout}. We additionally cover a range of general-purpose VLMs across scales: \textbf{Pixtral-12B} \citep{agrawal2024pixtral12b}, \textbf{MiMo-VL-7B} \citep{coreteam2025mimovltechnicalreport}, \textbf{Keye-VL-8B} \citep{kwaikeyeteam2025kwaikeyevltechnicalreport}, \textbf{Kimi-VL-16B-A3B} \citep{kimiteam2025kimivltechnicalreport}, and \textbf{Ovis2.5-9B} \citep{lu2025ovis25technicalreport}.

Finally, given the multilingual nature of \BENCHMARK, we include smaller and/or multilingual-oriented models: \textbf{Granite-Vision-3.3-2B} \citep{granitevisionteam2025granitevisionlightweightopensource}, \textbf{Idefics3-8B} \citep{laurencon2024buildingbetterunderstandingvisionlanguage}, and \textbf{EuroVLM} (1.7B, 9B) \citep{viveiros2025towervisionundermstandingimprovingmultilinguality}.

\section{Task Independence}
\label{app:task_independence}

\begin{figure}[htbp]
\centering
\includegraphics[width=\columnwidth]{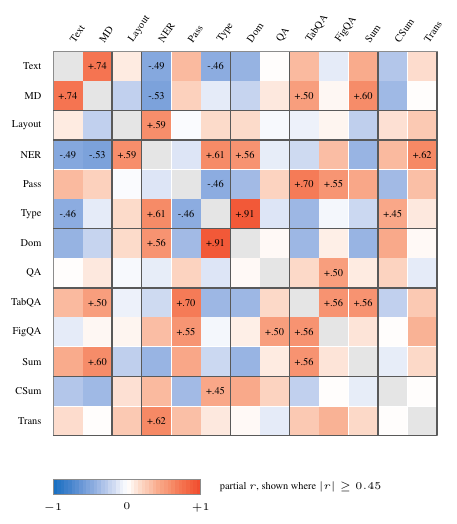}
\caption{Partial correlation between task scores across the \NMODELS\ evaluated models, controlling for each model's general ability (the mean over the remaining tasks). Red indicates capabilities that improve together beyond general ability; blue indicates trade-offs. Values shown where $|r|\geq0.45$. The mean off-diagonal partial correlation is $+0.09$, against $+0.66$ for the raw correlation: once model quality is removed, task performances are largely independent.}
\label{fig:task-corr}
\end{figure}

Figure~\ref{fig:task-corr} reports the partial correlation between task scores across the \NMODELS\ evaluated models, controlling for each model's general ability. The mean off-diagonal partial correlation is $+0.09$, against $+0.66$ for the raw correlation: once model quality is removed, performance on one task says little about performance on another. The correlations that survive fall within capability families rather than across them, which is what the suite is meant to separate: the $7$ within-family pairs average $+0.38$ against $+0.07$ for the $71$ cross-family pairs, a difference of $+0.32$ that a one-sided permutation test over $20{,}000$ relabellings of the $78$ pairs places at $p=0.009$.

\section{Complete Results}
\label{apx:full_results}

Table~\ref{tab:results} in the main text shows one model per family, the best-scoring one, so that the comparison fits on a page. This appendix gives the per-task scores for all \NMODELS\ evaluated systems with identical grouping, ordering and metrics, together with the per-language averages, which the main text reports only in prose. No model is omitted from the evaluation; the main text selects which rows to display, not which systems to run.

\begin{table*}[htbp]
\centering\scriptsize\setlength{\tabcolsep}{3.5pt}
\rowcolors{2}{}{black!5}
\begin{tabular}{@{}lcccccccccccccc@{}}
\toprule
\multirow{2}{*}{\textbf{Model (params)}} & \multirow{2}{*}{\textbf{AVG}} & \multicolumn{2}{c}{\textbf{Parsing}} & \multirow{2}{*}{\textbf{Layout}} & \multicolumn{2}{c}{\textbf{Inf.\ Extr.}} & \multicolumn{2}{c}{\textbf{Classif.}} & \multicolumn{3}{c}{\textbf{Question Answering}} & \multicolumn{2}{c}{\textbf{Summarization}} & \multirow{2}{*}{\textbf{Trans}} \\
\cmidrule(lr){3-4}\cmidrule(lr){6-7}\cmidrule(lr){8-9}\cmidrule(lr){10-12}\cmidrule(lr){13-14}
 &  & \textbf{Text} & \textbf{MD} &  & \textbf{NER} & \textbf{Pass} & \textbf{Type} & \textbf{Dom} & \textbf{QA} & \textbf{TabQA} & \textbf{FigQA} & \textbf{Sum} & \textbf{CSum} &  \\
\midrule
\rowcolor{white}\multicolumn{15}{@{}l}{\textbf{\textit{Small ($<$10B)}}} \\[1pt]
EuroVLM (1.7B) & 25.5 & 15.0 & 9.1 & 0.2 & 6.0 & 34.2 & 10.4 & 14.6 & 45.9 & 40.0 & 37.9 & 34.6 & 68.0 & 15.3 \\
EuroVLM (9B) & 40.6 & 24.6 & 41.5 & 11.6 & 20.5 & 39.5 & 51.1 & 58.7 & 53.9 & 44.8 & 42.5 & 40.2 & 77.0 & 21.4 \\
GLM-4.1V (9B) & 46.1 & 56.9 & 50.8 & 37.2 & 32.6 & 45.7 & 52.5 & 46.3 & 36.7 & 53.4 & 44.6 & 42.8 & 59.4 & 39.9 \\
Gemma3 (4B) & 48.5 & 51.3 & 37.1 & 26.0 & 30.5 & 43.0 & 57.1 & 67.8 & 48.8 & 53.7 & 47.4 & 39.1 & 88.0 & 41.3 \\
Gemma4 (E2B) & 50.6 & 56.8 & 58.0 & 28.3 & 30.0 & 40.7 & 59.7 & 60.0 & 50.0 & 56.5 & 49.3 & 44.1 & \textbf{90.5} & 34.3 \\
Gemma4 (E4B) & 58.0 & 68.0 & 69.6 & 41.3 & 34.1 & 42.8 & 69.5 & 74.7 & 53.7 & 63.8 & 52.8 & 45.9 & \underline{90.2} & 47.7 \\
Granite-Vision-3.3 (2B) & 35.4 & 52.9 & 10.6 & 41.9 & 19.1 & 35.9 & 23.5 & 39.7 & 51.1 & 31.5 & 38.0 & 24.9 & 68.1 & 23.4 \\
Idefics3 (8B) & 35.5 & 31.6 & 41.2 & 8.2 & 19.4 & 32.2 & 42.8 & 54.1 & 47.9 & 32.3 & 34.3 & 29.2 & 71.6 & 17.2 \\
InternVL3.5 (2B) & 51.5 & 68.6 & 69.9 & 25.2 & 24.4 & 40.6 & 58.9 & 59.5 & 51.7 & 60.5 & 47.0 & 47.0 & 76.4 & 40.0 \\
InternVL3.5 (4B) & 59.2 & 73.9 & 76.2 & \textbf{44.3} & 34.2 & 46.6 & 66.3 & 68.4 & 62.0 & 67.0 & 52.2 & 49.3 & 83.7 & 45.5 \\
InternVL3.5 (8B) & 58.9 & 74.2 & 75.6 & 43.0 & 36.9 & 44.7 & 62.6 & 70.3 & 59.7 & 63.9 & 50.6 & 49.3 & 85.6 & 49.3 \\
Keye-VL (8B) & 50.9 & 53.6 & 50.5 & 42.0 & 31.2 & 46.0 & 60.7 & 67.3 & 54.6 & 58.6 & 51.7 & 48.8 & 59.4 & 37.7 \\
MiMo-VL (7B) & 52.7 & 65.2 & 76.4 & 28.0 & 36.2 & 39.2 & 71.6 & 75.8 & 38.8 & 57.4 & 44.1 & 42.1 & 64.0 & 46.6 \\
Nanonets-OCR-s (3B) & 49.9 & \textbf{81.8} & 81.5 & 22.1 & 19.7 & 43.0 & 50.1 & 61.0 & 53.8 & 53.9 & 46.3 & 45.4 & 58.0 & 32.8 \\
Ovis2.5 (9B) & 47.4 & 49.2 & 51.4 & 30.3 & 29.6 & 40.3 & 60.6 & 71.1 & 35.9 & 54.9 & 47.6 & 40.7 & 65.0 & 40.0 \\
Qwen2-VL (2B) & 35.3 & 43.9 & 31.1 & 5.7 & 11.4 & 39.0 & 28.3 & 40.1 & 50.3 & 44.9 & 40.0 & 35.5 & 64.9 & 24.1 \\
Qwen2.5-VL (7B) & 59.3 & 73.9 & 78.0 & 26.1 & 33.1 & 47.2 & 68.0 & 74.4 & 60.9 & 62.6 & 55.4 & \underline{52.9} & 83.6 & 54.8 \\
Qwen3-VL (2B) & 48.1 & 60.0 & 70.3 & 24.3 & 22.2 & 37.3 & 51.1 & 61.0 & 53.0 & 53.4 & 49.6 & 40.0 & 76.1 & 26.7 \\
Qwen3-VL (4B) & 57.2 & 68.6 & 72.8 & 37.4 & 34.4 & 45.1 & 58.5 & 69.7 & 60.3 & 69.7 & 58.0 & 48.7 & 80.1 & 39.6 \\
Qwen3-VL (8B) & \underline{63.7} & 74.3 & \underline{82.5} & \underline{43.9} & 41.1 & \textbf{51.1} & 66.5 & 70.8 & \textbf{64.7} & \underline{71.5} & 62.4 & \textbf{53.1} & 87.8 & \underline{57.8} \\
Qwen3.5 (2B) & 55.4 & 72.5 & 78.6 & 22.9 & 29.1 & 48.5 & 52.7 & 64.3 & 61.0 & 66.3 & 58.6 & 48.4 & 73.8 & 43.5 \\
Qwen3.5 (4B) & 63.4 & 77.1 & 80.1 & 39.0 & \underline{45.4} & 47.8 & \underline{73.9} & \underline{75.9} & 60.5 & \textbf{71.9} & \underline{62.8} & 51.5 & 79.5 & \textbf{58.3} \\
Qwen3.5 (9B) & \textbf{64.2} & 75.9 & \textbf{83.9} & 42.5 & \textbf{46.2} & 49.0 & \textbf{74.3} & \textbf{77.9} & \underline{62.8} & 70.3 & \textbf{63.4} & 51.7 & 79.3 & 57.3 \\
dots.ocr (1.7B) & 35.0 & 75.6 & 73.6 & 6.5 & 0.3 & 43.7 & 0.9 & 0.7 & 39.6 & 57.5 & 35.0 & 45.0 & 51.3 & 25.2 \\
olmOCR-2 (7B) & 56.1 & \underline{81.5} & 79.9 & 24.9 & 24.2 & \underline{49.5} & 64.5 & 72.3 & 59.8 & 63.8 & 53.4 & 51.2 & 65.2 & 38.3 \\
\addlinespace[2pt]
\rowcolor{white}\multicolumn{15}{@{}l}{\textbf{\textit{Medium (10--40B)}}} \\[1pt]
Gemma3 (12B) & 53.7 & 62.0 & 41.7 & 30.3 & 39.2 & 44.0 & 61.8 & 75.0 & 49.3 & 58.6 & 51.9 & 42.9 & \textbf{90.9} & 50.3 \\
Gemma3 (27B) & 57.6 & 63.6 & 44.7 & 40.1 & 43.1 & 46.7 & 70.0 & 76.0 & 55.3 & 61.8 & 54.0 & 48.8 & 90.3 & 53.9 \\
Gemma4 (26B-A4B) & 60.7 & 62.7 & 59.4 & 48.8 & 44.4 & 42.1 & 80.8 & \textbf{84.4} & 54.1 & 61.1 & 55.7 & 50.0 & \underline{90.4} & 55.1 \\
Gemma4 (31B) & 65.4 & 73.2 & 70.3 & 52.7 & 52.4 & 43.6 & \underline{83.7} & 83.4 & 59.4 & 63.4 & 59.7 & \underline{53.2} & 89.4 & 66.2 \\
InternVL3.5 (14B) & 60.8 & 72.5 & 76.2 & 43.3 & 39.2 & 44.7 & 75.8 & 75.3 & 61.0 & 63.8 & 51.5 & 50.6 & 85.6 & 50.5 \\
InternVL3.5 (30B-A3B) & 58.4 & 64.7 & 75.6 & 51.4 & 36.3 & 42.4 & 69.0 & 76.6 & 56.9 & 62.8 & 50.6 & 48.9 & 86.3 & 37.6 \\
InternVL3.5 (38B) & 62.1 & 76.5 & 78.8 & 54.0 & 42.0 & 43.8 & 72.3 & 75.6 & 60.6 & 63.2 & 53.2 & 50.1 & 87.5 & 50.0 \\
Kimi-VL (16B-A3B) & 46.8 & 60.4 & 69.4 & 5.0 & 22.2 & 35.9 & 60.3 & 61.5 & 52.0 & 42.9 & 42.5 & 45.0 & 75.5 & 36.7 \\
Pixtral (12B) & 51.4 & 67.1 & 71.6 & 28.4 & 28.0 & 36.4 & 54.3 & 57.1 & 54.4 & 55.5 & 45.6 & 48.0 & 77.2 & 44.3 \\
Qwen2.5-VL (32B) & 62.5 & 66.8 & 78.4 & 48.7 & 46.9 & 47.9 & 70.1 & 77.6 & \underline{65.5} & \underline{75.7} & 61.8 & 47.7 & 62.5 & 63.1 \\
Qwen3-VL (30B-A3B) & 62.7 & \underline{80.0} & 83.8 & 39.8 & 43.3 & 45.7 & 74.4 & 77.4 & 57.9 & 64.0 & 58.8 & 50.5 & 83.6 & 55.9 \\
Qwen3-VL (32B) & 65.2 & 65.5 & 79.6 & 37.6 & 46.3 & \textbf{52.3} & 78.2 & 82.6 & \textbf{66.6} & \textbf{76.1} & \underline{66.3} & \textbf{53.3} & 78.8 & 64.3 \\
Qwen3.5 (27B) & \underline{68.1} & 76.8 & \underline{84.9} & \textbf{61.6} & \textbf{53.5} & 46.5 & 80.8 & \underline{84.1} & 62.2 & 68.8 & 63.9 & 53.1 & 82.2 & \underline{67.1} \\
Qwen3.5 (35B-A3B) & 66.0 & 77.0 & 84.4 & 41.6 & 51.7 & \underline{49.7} & 78.1 & 78.6 & 63.4 & 73.6 & \textbf{66.3} & 52.6 & 76.7 & 64.3 \\
Qwen3.6 (27B) & \textbf{69.0} & \textbf{80.6} & \textbf{88.5} & \underline{61.1} & 52.3 & 48.4 & \textbf{83.8} & 82.4 & 64.4 & 70.7 & 64.9 & 52.9 & 78.6 & \textbf{68.3} \\
Qwen3.6 (35B-A3B) & 65.9 & 76.9 & 84.5 & 46.9 & \underline{52.7} & 49.0 & 76.9 & 80.4 & 63.5 & 71.4 & 64.0 & 52.2 & 76.8 & 61.0 \\
\addlinespace[2pt]
\rowcolor{white}\multicolumn{15}{@{}l}{\textbf{\textit{Large ($>$40B)}}} \\[1pt]
GLM-4.5V (106B-A12B) & 59.2 & \underline{76.8} & 77.6 & \underline{44.0} & 41.8 & 48.4 & 60.1 & 67.1 & 49.5 & 63.2 & 55.2 & \underline{53.6} & 68.5 & \underline{63.4} \\
InternVL3 (78B) & 55.8 & 54.3 & 53.9 & 29.4 & 41.0 & 40.2 & 75.4 & 76.8 & 56.6 & 61.3 & 54.0 & 47.8 & \underline{87.9} & 47.0 \\
Qwen2.5-VL (72B) & \textbf{66.8} & \textbf{82.2} & \textbf{81.9} & \textbf{52.1} & \textbf{48.4} & \underline{48.7} & \underline{75.7} & \underline{79.6} & \underline{63.0} & \underline{67.3} & \underline{59.2} & \textbf{56.1} & \textbf{91.4} & 63.4 \\
Qwen3-VL (235B-A22B) & \underline{66.3} & 76.8 & \underline{80.0} & 38.6 & \underline{47.7} & \textbf{52.6} & \textbf{75.9} & \textbf{81.5} & \textbf{68.1} & \textbf{76.3} & \textbf{65.4} & 53.1 & 78.4 & \textbf{67.6} \\
\midrule
Gemini 3.8 Flash (commercial, API) & 65.8 & 73.8 & 71.1 & 44.5 & 50.9 & 46.0 & 79.5 & 79.0 & 59.3 & 74.1 & 65.2 & 53.3 & 87.0 & 71.7 \\
\bottomrule
\end{tabular}
\caption{Per-task scores on \BENCHMARK\ for all \NMODELS\ evaluated open-weight systems and the commercial reference: the complete version of Table~\ref{tab:results}. Grouping, ordering, metrics and emphasis follow that table. All scores $\times100$.}
\label{tab:results-full}
\end{table*}

\begin{table}[htbp]
\centering\footnotesize\setlength{\tabcolsep}{2.5pt}
\rowcolors{2}{}{black!5}
\resizebox{\columnwidth}{!}{%
\begin{tabular}{@{}lccccccc@{}}
\toprule
\multirow{2}{*}{\textbf{Model (params)}} & \multirow{2}{*}{\textbf{AVG}$^{*}$} & \multicolumn{5}{c}{\textbf{Language}} & \multirow{2}{*}{$\Delta$} \\
\cmidrule(lr){3-7}
 &  & \textbf{de} & \textbf{en} & \textbf{es} & \textbf{fr} & \textbf{it} &  \\
\midrule
\rowcolor{white}\multicolumn{8}{@{}l}{\textbf{\textit{Small ($<$10B)}}} \\[1pt]
EuroVLM (1.7B) & 27.8 & 25.8 & 31.3 & 28.4 & 26.7 & 27.0 & 5.5 \\
EuroVLM (9B) & 38.0 & 34.9 & 44.0 & 36.7 & 36.9 & 37.2 & 9.0 \\
GLM-4.1V (9B) & 45.5 & 42.4 & 46.8 & 46.7 & 44.7 & 46.6 & 4.4 \\
Gemma3 (4B) & 46.0 & 44.3 & 49.1 & 46.8 & 43.8 & 46.0 & 5.3 \\
Gemma4 (E2B) & 49.0 & 46.4 & 49.9 & 50.2 & 49.4 & 49.0 & 3.8 \\
Gemma4 (E4B) & 55.4 & 52.5 & 56.3 & 57.2 & 55.0 & 56.2 & 4.8 \\
Granite-Vision-3.3 (2B) & 36.1 & 33.4 & 41.0 & 36.5 & 35.8 & 34.1 & 7.6 \\
Idefics3 (8B) & 33.2 & 30.1 & 36.3 & 33.7 & 33.9 & 32.1 & 6.2 \\
InternVL3.5 (2B) & 50.1 & 47.9 & 51.3 & 51.3 & 50.0 & 50.1 & 3.4 \\
InternVL3.5 (4B) & 57.7 & 54.6 & 60.8 & 58.2 & 57.3 & 57.6 & 6.2 \\
InternVL3.5 (8B) & 57.5 & 55.1 & 61.3 & 57.6 & 56.6 & 57.1 & 6.1 \\
Keye-VL (8B) & 48.6 & 42.6 & 53.1 & 49.4 & 48.2 & 49.5 & 10.5 \\
MiMo-VL (7B) & 48.9 & 45.3 & 53.5 & 49.7 & 47.9 & 48.1 & 8.2 \\
Nanonets-OCR-s (3B) & 48.9 & 47.3 & 49.2 & 49.8 & 49.4 & 49.0 & \textbf{2.5} \\
Ovis2.5 (9B) & 44.1 & 40.7 & 47.7 & 44.4 & 44.0 & 43.6 & 7.0 \\
Qwen2-VL (2B) & 35.5 & 33.7 & 36.1 & 36.2 & 35.2 & 36.4 & \underline{2.8} \\
Qwen2.5-VL (7B) & 57.1 & 55.2 & 58.2 & 57.7 & 56.6 & 57.9 & 2.9 \\
Qwen3-VL (2B) & 46.6 & 43.4 & 50.4 & 47.9 & 46.3 & 45.1 & 7.0 \\
Qwen3-VL (4B) & 55.9 & 52.8 & 58.3 & 56.8 & 55.3 & 56.1 & 5.5 \\
Qwen3-VL (8B) & \textbf{62.8} & \textbf{60.0} & \textbf{65.9} & 62.9 & \textbf{61.8} & \textbf{63.3} & 5.9 \\
Qwen3.5 (2B) & 54.8 & 51.7 & 59.3 & 55.3 & 53.7 & 54.3 & 7.6 \\
Qwen3.5 (4B) & 61.3 & 58.4 & \underline{64.4} & \underline{63.0} & 59.6 & 60.8 & 6.0 \\
Qwen3.5 (9B) & \underline{62.0} & \underline{59.4} & 63.6 & \textbf{63.7} & \underline{61.3} & \underline{62.2} & 4.3 \\
dots.ocr (1.7B) & 41.2 & 38.5 & 44.9 & 40.6 & 40.9 & 41.2 & 6.4 \\
olmOCR-2 (7B) & 53.8 & 52.5 & 55.7 & 54.0 & 52.3 & 54.5 & 3.4 \\
\addlinespace[2pt]
\rowcolor{white}\multicolumn{8}{@{}l}{\textbf{\textit{Medium (10--40B)}}} \\[1pt]
Gemma3 (12B) & 51.0 & 48.3 & 53.8 & 52.0 & 49.0 & 51.8 & 5.5 \\
Gemma3 (27B) & 54.7 & 53.2 & 55.5 & 55.0 & 53.3 & 56.7 & 3.5 \\
Gemma4 (26B-A4B) & 56.7 & 53.8 & 56.4 & 58.5 & 57.0 & 57.9 & 4.8 \\
Gemma4 (31B) & 62.1 & 59.2 & 63.3 & 63.3 & 61.7 & 63.2 & 4.1 \\
InternVL3.5 (14B) & 58.1 & 56.2 & 60.6 & 57.9 & 57.8 & 57.8 & 4.4 \\
InternVL3.5 (30B-A3B) & 55.8 & 53.2 & 59.1 & 53.9 & 56.3 & 56.3 & 5.9 \\
InternVL3.5 (38B) & 60.0 & 57.4 & 62.3 & 60.7 & 59.7 & 59.9 & 4.9 \\
Kimi-VL (16B-A3B) & 44.3 & 42.1 & 48.1 & 43.4 & 43.9 & 44.1 & 6.0 \\
Pixtral (12B) & 50.6 & 46.9 & 55.7 & 50.6 & 50.3 & 49.5 & 8.8 \\
Qwen2.5-VL (32B) & 60.4 & 58.1 & 62.5 & 61.1 & 58.8 & 61.7 & 4.4 \\
Qwen3-VL (30B-A3B) & 60.3 & 57.7 & 64.5 & 60.3 & 59.4 & 59.5 & 6.8 \\
Qwen3-VL (32B) & 62.4 & 59.7 & 64.5 & 63.7 & 60.7 & 63.4 & 4.8 \\
Qwen3.5 (27B) & \underline{65.5} & \underline{63.5} & \underline{67.3} & \underline{65.6} & \underline{64.8} & \underline{66.2} & 3.8 \\
Qwen3.5 (35B-A3B) & 63.8 & 62.0 & 65.1 & 64.9 & 62.2 & 64.6 & \underline{3.0} \\
Qwen3.6 (27B) & \textbf{66.4} & \textbf{65.4} & \textbf{67.4} & \textbf{67.4} & \textbf{65.2} & \textbf{66.6} & \textbf{2.2} \\
Qwen3.6 (35B-A3B) & 63.5 & 61.2 & 65.2 & 64.4 & 62.4 & 64.5 & 4.0 \\
\addlinespace[2pt]
\rowcolor{white}\multicolumn{8}{@{}l}{\textbf{\textit{Large ($>$40B)}}} \\[1pt]
GLM-4.5V (106B-A12B) & 58.4 & 55.0 & 60.1 & 60.0 & 58.0 & 58.8 & \underline{5.1} \\
InternVL3 (78B) & 52.1 & 49.0 & 55.9 & 51.7 & 52.7 & 51.3 & 6.9 \\
Qwen2.5-VL (72B) & \textbf{64.9} & \textbf{62.7} & \underline{65.6} & \textbf{66.1} & \textbf{64.0} & \textbf{65.9} & \textbf{3.4} \\
Qwen3-VL (235B-A22B) & \underline{64.1} & \underline{61.1} & \textbf{66.5} & \underline{65.6} & \underline{63.0} & \underline{64.1} & 5.3 \\
\bottomrule
\end{tabular}}

\caption{Headline average per language for all \NMODELS\ evaluated systems, grouped and ordered as in Table~\ref{tab:results}. $\Delta$ is the spread between a model's best and worst language. Best per column within each size group in bold, runner-up underlined (for $\Delta$, lowest is best). AVG$^{*}$ is computed over 11 of the \NTASKS\ tasks, excluding Type Classification and Domain Classification, which are underpowered; it is therefore not comparable with the AVG column of Table~\ref{tab:results}. All scores $\times100$.}
\label{tab:languages-full}
\end{table}

\section{Per-Domain Breakdown of the Classification Tasks}
\label{app:domain_classification}

\begin{figure}[htbp]
\centering
\includegraphics[width=\columnwidth]{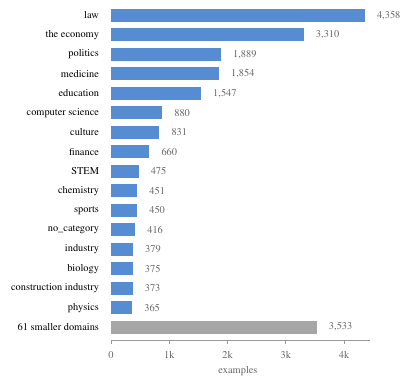}
\caption{Examples per domain, over the 22{,}146 examples to which the classifier assigns a domain. The 16 largest of the 77 domains are shown individually; the remaining 61 are pooled into the last bar.}
\label{fig:domain_distribution}
\end{figure}
Type and Domain Classification are the two tasks whose own labels are domains, which makes them the natural candidates for a per-domain breakdown, and we computed it. It is not reportable at this granularity, for two reasons that are worth stating rather than hiding. They are annotated on a coarser 15-label taxonomy that does not merge with the 77-domain tagging used above, so their rows would not be comparable with the others. And with 500 examples per task spread over its label set, the median 95\% confidence interval of a per-domain cell is $\pm12.7$ points against a total spread of $20.3$: the interval is nearly as wide as the effect. The ordering we obtain (\emph{computer science} and \emph{natural sciences} easiest, \emph{technology}, \emph{culture} and \emph{politics} hardest, for both tasks) is stable across the two tasks and therefore probably real, but we do not report per-domain numbers for them, because a table of 28 cells at that precision would imply resolution the sample size does not support.

\fi %

\end{document}